\documentclass{article} % For LaTeX2e
\usepackage{iclr2027_conference,times}
\iclrfinalcopy
\usepackage{amsmath,amsfonts,bm}

\def\eqref#1{equation~\ref{#1}}
\def\1{\bm{1}}

\DeclareMathAlphabet{\mathsfit}{\encodingdefault}{\sfdefault}{m}{sl}
\SetMathAlphabet{\mathsfit}{bold}{\encodingdefault}{\sfdefault}{bx}{n}

\usepackage{hyperref}
\usepackage{enumitem}
\usepackage{url}
\usepackage{booktabs}
\usepackage{algorithmic}
\usepackage{graphicx}
\usepackage{textcomp}
\usepackage{xcolor}
\usepackage{graphicx}
\usepackage{subcaption}
\usepackage[most]{tcolorbox}
\usepackage{tabularx}
\usepackage{xcolor}
\usepackage{wrapfig}
\usepackage[T1]{fontenc}
\usepackage{listings} 
\usepackage{multirow}

\makeatletter
\newcommand{\FUNCTION}[1]{\STATE \textbf{function} #1 \begin{ALC@g}}
\newcommand{\ENDFUNCTION}{\end{ALC@g}\STATE \textbf{end function}}
\makeatother
\usepackage{tikz}
\usetikzlibrary{tikzmark,decorations.pathreplacing,calc}

\usepackage{algorithm}
\usepackage{algorithmic}
\usepackage{amsmath}
\usepackage{amssymb}

\title{HGM-Dual: Lineage-Aware Guided Agent Harness Optimization for Long-Horizon Planning}

\author{
Sudipta Paul\thanks{Equal contribution.} \And
Pengrui Quan\footnotemark[1] \And
Aounon Kumar \And
Nicholas Tzou \And
Vivek Kulkarni \And
Srinivas Chappidi
\\[0.7em]
\makebox[\textwidth][c]{Samsung Research America}
\\
\makebox[\textwidth][c]{\small
\texttt{\{sudipta.paul, pengrui.quan, aounon.kumar, n.tzou,\}}}
\\[-0.2em]
\makebox[\textwidth][c]{\small
\texttt{\{v.kulkarni1, vasu.c\}@samsung.com}}
}

\begin{document}

\maketitle

\begin{abstract}
Harness optimization, which improves an agent's prompts, tools, and
workflows, is an increasingly effective way to enhance agents built on frozen or closed-source LLMs. Leading approaches such as the Darwin--G\"odel Machine and the Huxley--G\"odel Machine (HGM) rely on recursive self-improvement and were developed mainly for coding, where the competence needed to improve the agent coincides with the competence being improved. Planning tasks, in which
an agent must produce a sequence of tool calls that jointly satisfies interacting constraints, break this alignment. There, open-ended self-modification repeatedly explores ineffective changes and consumes substantial evaluation budgets. We introduce HGM-Dual, a diagnostic-guided framework for budget-efficient harness evolution for planning agents. First, HGM-Dual represents agent evolution as lineage-aware context. This context records what each ancestor changed, which behaviors improved or regressed, and which modifications were ineffective, giving the editor explicit edit–outcome attribution instead of an unstructured archive of past agents. Second, HGM-Dual introduces error-conditioned dual-path child generation. For each selected parent, it generates a general child that preserves open-ended exploration and specialized children that target the harness components most likely responsible for the parent's highest-ranked constraint violations. HGM-Dual outperforms the strongest baseline by up to $14.45\%$ relative improvement in case accuracy while matching HGM variants' final performance with $2\times$--$4\times$ fewer evaluation budget.
\end{abstract}

\section{Introduction}
\label{sec:introduction}

Agent harness optimization~\cite{khattab2024dspy, yuksekgonul2024textgrad} improves large language model (LLM) agents without updating model parameters by optimizing the prompts, tools, and workflows that govern agent behavior. Recent self-improving systems \cite{novikov2025alphaevolve} automate this process through evolutionary search. The Darwin--G\"odel Machine (DGM)~\cite{zhang2026darwin} maintains an archive of agents and generates new candidates by modifying selected parents, while the Huxley--G\"odel Machine (HGM)~\cite{wang2026huxley} selects candidates based on \emph{clade-metaproductivity}, their potential to produce strong descendants.

This paradigm is well suited to coding agents because the ability being improved, understanding and modifying code, also enables self-modification. Planning tasks, however, require sequences of actions that satisfy interacting constraints~\cite{zhang2026deepplanning, xie2024travelplanner}, creating a mismatch that poses two challenges for existing recursive self-improvement methods.

\begin{wrapfigure}{r}{0.58\textwidth}
    \centering
    \includegraphics[width=\linewidth]{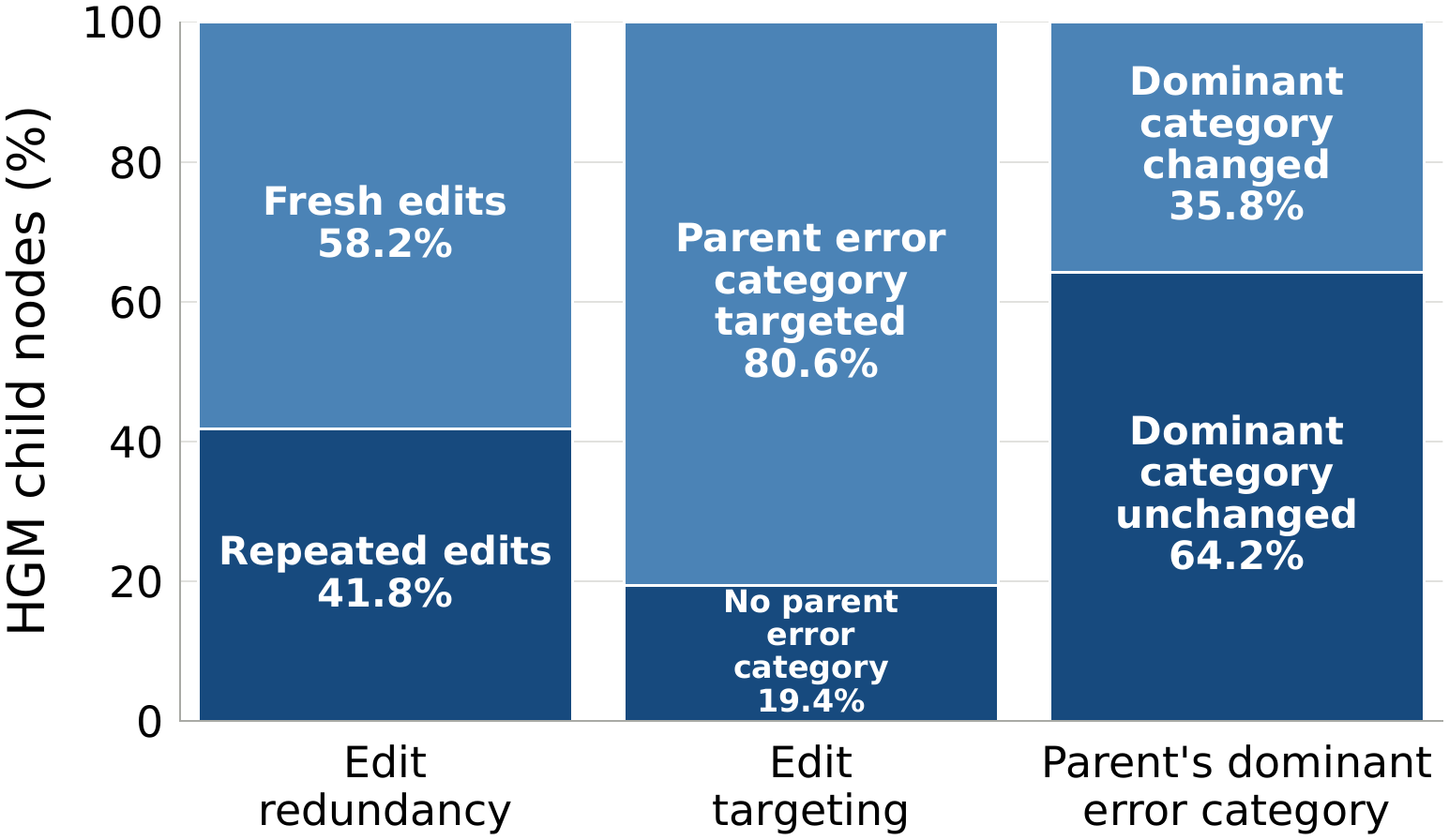}
    \caption{Edit behavior in HGM. Among child nodes,
    41.8\% repeat previous edits, 19.4\% target an error
    category absent from the parent's observed failures,
    and 64.2\% retain the parent's dominant error category.
    See Appendix~\ref{sec:edit_redundancy} for details.}
    \label{fig:edit-analysis}
\end{wrapfigure}

First, \emph{effective improvement requires a structured context of the search
history}. Planning benchmarks expose a relatively small and recurring set of
constraint failures, so different agents in the search tree often encounter
similar problems. If an editor observes only its current parent's evaluation logs, it cannot determine whether a similar fix has already
been attempted elsewhere, whether it improved or degraded performance, or
which previous modifications led to the current behavior. As a result,
different branches may repeatedly rediscover the same ineffective changes.
In HGM's shopping search, 41.8\% of edits repeat an edit pattern already
present in the tree (Figure~\ref{fig:edit-analysis}, left), with each repeated
attempt consuming another full planning evaluation. What is needed is not
simply access to more previous agents, but explicit \emph{edit--outcome
context}: a concise record of what was changed, what behavior changed as a
result, and which changes succeeded, regressed, or proved ineffective.

\vspace{-5pt}

Second, \emph{improvement must be targeted at the specific constraints the
agent violates}. Planning success depends on the correctness of many
interdependent decisions: a single infeasible transfer, invalid coupon, or
unmet product requirement can invalidate an otherwise valid itinerary or
cart. Consequently, a broad modification that improves behavior in general may
still leave the failing case unresolved unless it corrects the particular
constraint violation responsible for failure. Without explicit guidance,
open-ended editing may instead modify an unrelated harness component or make
a diffuse change that does not address the observed error. In HGM's shopping
search, 19.4\% of edits target a constraint category that the parent never
failed (Figure~\ref{fig:edit-analysis}, middle), and in 64.2\% of edits the
child retains its parent's dominant failure category
(Figure~\ref{fig:edit-analysis}, right). These results suggest that efficient
self-improvement for planning requires both learning from prior edit outcomes
and directing new edits toward the specific constraint failures observed in
the current agent.

We introduce the \emph{Huxley--G\"odel Machine Dual} (HGM-Dual), a
diagnostic-guided framework for budget-efficient agent harness evolution. First, it represents
agentic evolution as structured contextual information. This \emph{agent-evolution context} records what was changed, which
behaviors improved or regressed, and which attempted modifications were
ineffective. The editor therefore receives explicit edit--outcome attribution
instead of treating previous agent snapshots as an unstructured archive. This
context discourages repeated failures and allows subsequent edits to build
systematically on evidence collected throughout the search.

Second, HGM-Dual introduces \emph{error-conditioned dual-path child generation}. For each selected parent, the framework generates complementary children: a \emph{general child}, which preserves the broad and open-ended exploration of existing evolutionary methods, and \emph{specialized children}, which target the harness component most likely responsible for the parent's observed errors. The general path retains the ability to discover unexpected improvements, while the specialized path provides diagnostic-guided local refinement. Together, the dual paths balance exploration with targeted exploitation and direct the evaluation budget toward more informative regions of the harness search space.

Experiments on travel and shopping planning tasks show that HGM-Dual outperforms the strongest baseline by up to $14.45\%$ relative improvement in case accuracy while matching HGM variants' final performance with $2\times$--$4\times$ fewer evaluation budget. These results demonstrate that structured evolutionary context and diagnostic guidance can make harness self-improvement both more effective and substantially more resource-efficient.
\section{Related Works}
\noindent\textbf{Harness Optimization through Self-Improvement.}
Recent methods improve frozen LLMs by automatically modifying their agentic harnesses, including prompts, tools, skills, and workflows \cite{khattab2024dspy, opsahl2024optimizing, zhang2025aflow, wang2023voyager, chen2026avo, karten2026continual, lou2026autoharness, zhang2026self}. ADAS~\cite{hu2025automated}, DGM~\cite{zhang2026darwin}, HGM~\cite{wang2026huxley}, GEPA~\cite{agrawal2026gepa}, and HyperAgents~\cite{zhang2026hyperagents} search over agent implementations through iterative or evolutionary self-modification. Meta-Harness~\cite{lee2026meta} gives an agentic proposer access to the code, scores, and execution traces of prior candidates, while SkillOpt~\cite{yang2026skillopt} optimizes a natural-language skill through bounded edits and validation. Collectively, these methods demonstrate that harness optimization can substantially improve agent performance without updating model parameters.

Nevertheless, two challenges remain. First, prior edit experience is either local or excessive: DGM~\cite{zhang2026darwin} and HGM~\cite{wang2026huxley} primarily generate modifications from the selected parent's code and evaluation logs, without providing enough context on the effectiveness of all edits attempted. On the other hand, archive-based methods~\cite{hu2025automated, lee2026meta, lin2026agentic, jiang2026deltaevolve} may flood the evolution context with numerous previous candidates and raw execution traces. This can obscure useful edit-outcome relationships and lead to repeated ineffective modifications across the search tree. Second, although prior methods guide candidate selection or validation~\cite{yang2026skillopt, wang2026huxley, zhang2026darwin, zhang2026hyperagents}, edit generation itself remains largely open-ended and may not target the harness component responsible for observed failures. Such weakly targeted exploration can require substantial evaluation resources. HGM-Dual addresses these limitations by maintaining a concise, structured agent-evolution context and by generating both a general child for broad exploration and an error-conditioned specialized child for targeted refinement.

% \noindent\textbf{Constrainted Long-Horizon Planning.}
% We study the challenging travel and shopping tasks in
% DeepPlanning~\cite{zhang2026deepplanning}. The challenge is to maintain local and global consistency across the entire trajectory while satisfying the task constraints. Travel planning requires minute-level itineraries of up to seven days using nine APIs and is evaluated by 21 feasibility checkpoints across eight dimensions, including temporal, spatial, and budget consistency. Shopping planning requires searching an average of 171 records through 15 APIs over a schema containing 45 product and user attributes. Each task imposes a different combination of product identity, quantity, price, brand, color, size, material, rating, stock, delivery, and user-profile constraints. The agent must satisfy these requirements while jointly selecting products and coupons under eligibility, threshold, scope, stacking, and budget rules to minimize the final cart price.
\section{HGM-Dual}
\subsection{Problem Setup}
\label{sec:problem_setup}

% \noindent\textbf{Agent Harness Optimization.}
% We define an agent harness as
% \begin{equation}
%     \mathcal{H}=(P,\mathcal{T},W),
% \end{equation}
% where \(P\), \(\mathcal{T}\), and \(W\) denote the prompt, tools, and agentic workflow, respectively. Given a frozen LLM, our goal is to optimize its harness:
% \begin{equation}
%     \mathcal{H}^{*}
%     =\arg\max_{\mathcal{H}}
%     \mathbb{E}_{x\sim\mathcal{D}}
%     \left[R\!\left(x,A_{\mathcal{H}}(x)\right)\right].
% \end{equation}

\noindent\textbf{Agent Harness Optimization.}
We consider a frozen language model whose behavior is controlled by an agent harness
\begin{equation}
\mathcal{H}=(\mathcal{P},\mathcal{T},\mathcal{W}),
\end{equation}
where $\mathcal{P}$ specifies the prompt and instructions provided to the model, $\mathcal{T}$ denotes the set of available tools, and $\mathcal{W}$ defines the agentic workflow that governs how the model and tools interact to solve a task. The parameters of the underlying language model remain fixed throughout optimization; only the components of $\mathcal{H}$ are modified.

Given a task $x$ sampled from a task distribution $\mathcal{D}$, executing the frozen language model under harness $\mathcal{H}$ produces an output $\mathcal{H}(x)$. We evaluate this output using a task-specific scoring function $J(x,\mathcal{H}(x))$, which measures the quality of the harness execution on task $x$. Agent harness optimization seeks a harness that maximizes expected performance over the task distribution:
\begin{equation}
\mathcal{H}^{*}
=
\arg\max_{\mathcal{H}}
\mathbb{E}_{x\sim\mathcal{D}}
\left[
J\!\left(x,\mathcal{H}(x)\right)
\right].
\label{eq:harness_optimization}
\end{equation}

\subsection{Overview}
\label{sec:hgm_dual_overview}

\begin{wrapfigure}{r}{0.40\textwidth}
\centering
\includegraphics[width=\linewidth]{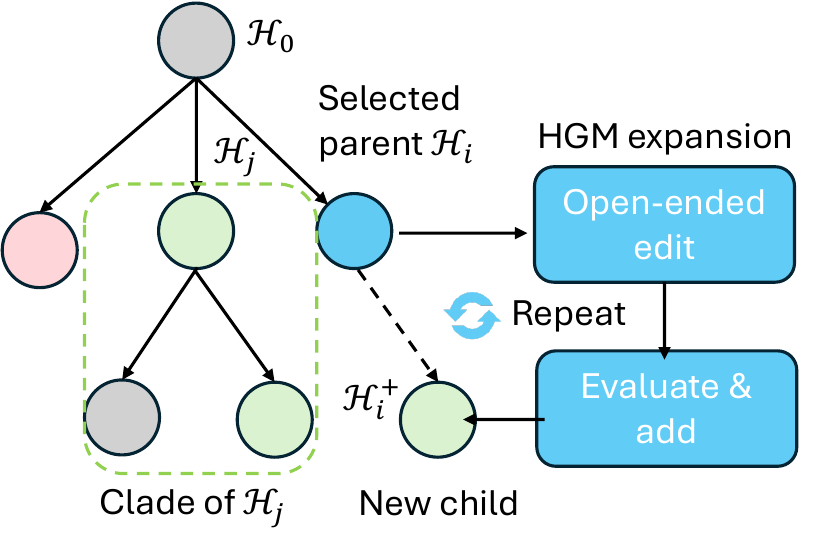}
\caption{HGM search: select a parent, generate an
open-ended edit, evaluate the child, and add it
to the archive.}
\label{fig:hgm-search}
\end{wrapfigure}
\vspace{-5pt}

\paragraph{HGM.} 
HGM optimizes an agent harness through
evolutionary self-modification. Starting from an initial harness
$\mathcal{H}_0$, HGM maintains an archive $\mathcal{A}$ of previously
evaluated harnesses organized as an evolutionary tree
(Figure~\ref{fig:hgm-search}). Each node corresponds to a harness, and
each edge corresponds to a modification that produces a child from its
parent. At each iteration, HGM selects a parent harness
$\mathcal{H}_i \in \mathcal{A}$ using its \emph{clade-metaproductivity}
criterion, which is designed to favor candidates with the potential to
produce strong descendants. An editor then examines the selected
harness and its evaluation feedback, proposes an open-ended
modification, evaluates the resulting child, and adds it to the
archive. The editor primarily conditions on the selected harness and its current evaluation context
$\varepsilon_i$:
\begin{equation}
\mathcal{H}'
=
\operatorname{Edit}
\left(
\mathcal{H}_i,
\varepsilon_i
\right).
\label{eq:hgm_edit}
\end{equation}
Thus, the archive records an evolutionary history, but that history is
not explicitly represented as structured edit-outcome evidence for
the editor. Moreover, because the edit is
open-ended, it is not explicitly constrained to address the failures
observed in the selected parent.

\begin{figure*}[t]
\centering
\includegraphics[width=.8\textwidth]{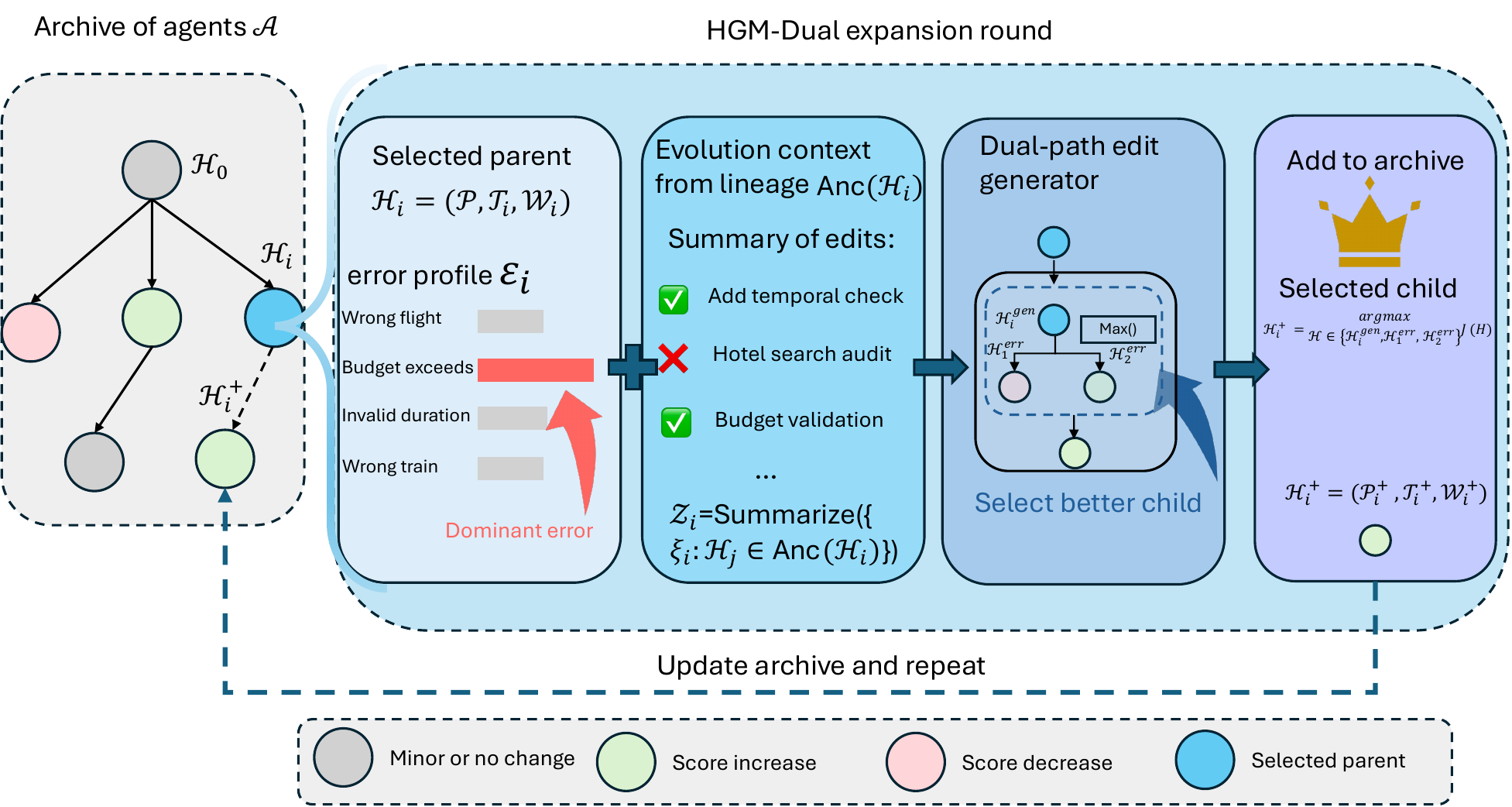}
\caption{Overview of HGM-Dual and one expansion round. Given a selected parent $\mathcal{H}_i$, HGM-Dual constructs
an evolution context $\mathcal{Z}_i$ by summarizing edits and outcomes
along its ancestral path $\mathrm{Anc}(\mathcal{H}_i)$. The evolution
context is provided to both editing paths: a general editor proposes
an open-ended improvement, while specialized editors target the
highest-ranked errors in the parent's error profile $\varepsilon_i$.
The resulting children are evaluated under the same protocol, and only
the strongest child is added to the archive.}
\label{fig:hgm-dual-explain}

\vspace{-5pt}
\end{figure*}

\paragraph{HGM-Dual.}

HGM-Dual retains HGM's evolutionary search procedure but changes how each selected parent is expanded (we refer readers to Section \ref{sec:full-hgm-dual} for full integration). As illustrated in
Figure~\ref{fig:hgm-dual-explain}, it introduces two mechanisms.
First, rather than conditioning an edit only on the selected harness
$\mathcal{H}_i$ and its evaluation context $\varepsilon_i$,
HGM-Dual additionally provides an evolution context $\mathcal{Z}_i$
that summarizes the edits and outcomes along its ancestral lineage:
\begin{equation}
\mathcal{H}'
=
\operatorname{Edit}
\left(
\mathcal{H}_i,
\varepsilon_i,
\mathcal{Z}_i
\right).
\label{eq:evolution_conditioned_edit}
\end{equation}

Second, HGM-Dual augments HGM's open-ended child generation with
error-conditioned editing. Each expansion generates both general and error-conditioned children. Thus, HGM-Dual retains HGM's evolutionary backbone while making edit generation both \emph{history-aware} and \emph{error-aware}. We summarize the HGM-Dual expansion process in Algorithm \ref{alg:hgm-dual-expansion} and describe
them in Sections~\ref{sec:agent_evolution_context}
and~\ref{sec:dual_path_generation}, respectively.

\subsection{Agent Evolution as Interpretable Context}
\label{sec:agent_evolution_context}

An editor observing only the current harness and its evaluation feedback cannot reliably attribute performance changes to prior edits or identify previously failed modifications. Providing the full archive and raw traces would address this limitation but is costly and difficult to interpret. HGM-Dual therefore compresses the relevant evolutionary history into a structured context.

For each non-root harness $\mathcal{H}_i$ with parent $\mathcal{H}_{p_i}$, we define an evolution record
\begin{equation}
\zeta_i =
\left(
d_i,\,
u_i,\,
\Delta \widehat{J}_i,\,
\Delta \mathbf{f}_i,\,
b_i
\right),
\label{eq:evolution_record}
\end{equation}
where $d_i$ summarizes the edit, $u_i$ records functionality added, removed, or invoked,
\begin{equation}
\Delta \widehat{J}_i
=
\widehat{J}(\mathcal{H}_i)
-
\widehat{J}(\mathcal{H}_{p_i}),
\label{eq:delta_score}
\end{equation}
measures its performance impact, and
\begin{equation}
\Delta \mathbf{f}_i
=
\mathbf{f}_i-\mathbf{f}_{p_i}
\label{eq:delta_error}
\end{equation}
captures changes in constraint-level errors. Finally, $b_i$ summarizes the observed behavioral consequences. Thus, $\zeta_i$ links each modification to its functional, performance, and behavioral effects.

Let $\operatorname{Anc}(\mathcal{H}_i)=(\mathcal{H}_0,\ldots,\mathcal{H}_{p_i},\mathcal{H}_i)$ denote the ordered ancestral path to $\mathcal{H}_i$. We define its lineage context as
\begin{equation}
\mathcal{Z}_i
=
\operatorname{Summarize}
\left(
\left\{
\zeta_j :
\mathcal{H}_j \in
\operatorname{Anc}(\mathcal{H}_i)
\right\}
\right).
\label{eq:lineage_context}
\end{equation}
This bounded summary preserves which functionality was introduced and used, which edits produced improvements or regressions, and how behavior evolved along the lineage, while consolidating redundant records.

When expanding $\mathcal{H}_i$, the editor receives its source, current evaluation context $\varepsilon_i$, and lineage context $\mathcal{Z}_i$. This compact memory helps avoid repeating failed changes, retain complementary improvements, and identify interactions between edits that are not apparent from the current score alone.

\begin{algorithm}[t]
\small
\caption{HGM-Dual Expansion}
\label{alg:hgm-dual-expansion}
\begin{algorithmic}[1]
\REQUIRE Archive $\mathcal{A}$ with evolution records
$\zeta_j = (d_j,u_j,\Delta J_j,\Delta\mathbf{e}_j,b_j)$;
selected parent $\mathcal{H}_i$ with evaluation context
$\varepsilon_i = (J_i,\mathbf{f}_i,\mathcal{S}_i)$;
error categories $\mathcal{C}$;
fitness $\widehat{J}$;
pre-search stability statistics $\kappa,\rho_B$.
$\operatorname{Summarize}$ compresses ancestral evolution records;
$\operatorname{Edit}$ proposes a new harness with evolution context.
\ENSURE Updated archive $\mathcal{A}$.

\FUNCTION{$\textsc{Expand}(\mathcal{H}_i, \varepsilon_i, \mathcal{A})$}

% ---------- Build evolution context ----------
\STATE
$\operatorname{Anc}(\mathcal{H}_i)
\leftarrow
(\mathcal{H}_0,\ldots,\mathcal{H}_{p_i},\mathcal{H}_i)$
\tikzmark{ctx-top}

\STATE
$\mathcal{Z}_i
\leftarrow
\operatorname{Summarize}
\left(
\{\zeta_j :
\mathcal{H}_j \in \operatorname{Anc}(\mathcal{H}_i)\}
\right)$
\tikzmark{ctx-bottom}

% ---------- Generic editing ----------
\STATE
$\mathcal{H}_i^{\mathrm{gen}}
\leftarrow
\operatorname{Edit}_{\mathrm{gen}}
(\mathcal{H}_i;\varepsilon_i,\mathcal{Z}_i)$
\tikzmark{dual-top}

% ---------- Stability-gated error conditioning ----------
\IF{$\kappa \geq 0.4 \ \land\ \rho_B \geq 0.7$}

    \STATE
    $\{c_{i,1}^{*},c_{i,2}^{*}\}
    \leftarrow
    \operatorname{Top2}_{c\in\mathcal{C}}
    (f_{i,c})$

    \FOR{$k \in \{1,2\}$}
        \STATE
        $\mathcal{H}_{i,k}^{\mathrm{err}}
        \leftarrow
        \operatorname{Edit}_{\mathrm{err}}
        \left(
        \mathcal{H}_i;
        c_{i,k}^{*},
        \mathcal{S}_i^{c_{i,k}^{*}},
        \mathcal{Z}_i
        \right)$
        \tikzmark{err-line}
    \ENDFOR

    \STATE
    $\mathcal{H}_i^{+}
    \leftarrow
    \displaystyle
    \arg\max_{\mathcal{H}\in
    \{
    \mathcal{H}_i^{\mathrm{gen}},
    \mathcal{H}_{i,1}^{\mathrm{err}},
    \mathcal{H}_{i,2}^{\mathrm{err}}
    \}}
    \widehat{J}(\mathcal{H})$

\ELSE

    \STATE
    $\mathcal{H}_i^{+}
    \leftarrow
    \mathcal{H}_i^{\mathrm{gen}}$
    \hfill $\triangleright$ skip error-conditioned generation

\ENDIF
\tikzmark{dual-bottom}

% ---------- Archive update ----------
\STATE
Construct $\varepsilon_i^{+}$ and
$\zeta_i^{+}
=
(d_i^{+},u_i^{+},
\Delta J_i^{+},
\Delta\mathbf{e}_i^{+},
b_i^{+})$
relative to $\mathcal{H}_i$

\STATE
$\mathcal{A}
\leftarrow
\mathcal{A}\cup\{\mathcal{H}_i^{+}\}$
\hfill
$\triangleright$ store with
$\mathcal{H}_i$, $\varepsilon_i^{+}$, $\zeta_i^{+}$

\RETURN $\mathcal{A}$
\ENDFUNCTION

% ---------- Braces (kept inside algorithmic: adds no extra line) ----------
\begin{tikzpicture}[overlay, remember picture]
\coordinate (CT) at (pic cs:ctx-top);
\coordinate (CB) at (pic cs:ctx-bottom);
\coordinate (DT) at (pic cs:dual-top);
\coordinate (EL) at (pic cs:err-line);
\coordinate (DB) at (pic cs:dual-bottom);

% x-position: just right of the longest line
\path let
    \p1 = (CT),
    \p2 = (CB),
    \p3 = (DT),
    \p4 = (EL)
in
coordinate (X) at ({max(\x1,\x2,\x3,\x4) + 0.55cm}, 0);

\coordinate (M) at ($(CB)!0.5!(DT) + (0, 0.12)$);

\draw[
    decorate,
    decoration={brace, amplitude=5pt},
    thick
]
($(CT -| X) + (0, 0.30)$) --
($(M -| X) + (0, 0.05)$)
node[midway, right=7pt, align=left]
{Build evolution\\context};

\draw[
    decorate,
    decoration={brace, amplitude=5pt},
    thick
]
($(M -| X) + (0, -0.05)$) --
($(DB -| X) + (0, -0.12)$)
node[midway, right=7pt, align=left]
{Dual-path generation};

\end{tikzpicture}

\end{algorithmic}
\end{algorithm}
\vspace{-7pt}

% \subsection{Error-Conditioned Dual-Path Child Generation}
% \label{sec:dual_path_generation}

% The space of possible modifications to $\mathcal P$, $\mathcal{T}$, and $\mathcal W$ is large, and unrestricted generation frequently produces plausible but irrelevant edits. HGM-Dual guides each archive expansion through two complementary editing paths. The general path supports broad exploration of the harness design, whereas the error-conditioned path concentrates computation on verified weaknesses of the selected harness.

\subsection{Error-Conditioned Dual-Path Child Generation}
\label{sec:dual_path_generation}

The space of possible modifications to $\mathcal P$, $\mathcal T$, and $\mathcal W$ is large, and unrestricted generation frequently produces plausible but irrelevant edits. HGM-Dual therefore uses two complementary editing paths: a general path for broad exploration and an error-conditioned path for targeting verified weaknesses. To avoid specializing to noisy failure signals, the error-conditioned path is enabled only when a one-time pre-search test shows sufficient task-level agreement ($\kappa \geq 0.4$) and dominant-category stability ($\rho_B \geq 0.7$); otherwise, expansion uses only the general path.

For a selected harness $\mathcal{H}_i$, let
\begin{equation}
\varepsilon_i
=
\left(
J(\mathcal{H}_i),
\mathbf{f}_i,
\mathcal{S}_i
\right),
\qquad
\mathbf{f}_i
=
\left(f_{i,c}\right)_{c\in\mathcal{C}},
\end{equation}
denote its evaluation context, where $J(\mathcal{H}_i)$ is its evaluation score, $\mathbf{f}_i$ is the vector of
evaluator-category error frequencies,
$\mathcal{C}$ is the set of evaluator-defined error categories, and
$\mathcal{S}_i$ contains representative execution traces and evaluator
feedback. The general child is produced by
\begin{equation}
\mathcal{H}^{\mathrm{gen}}_i
=
\operatorname{Edit}_{\mathrm{gen}}
\left(
\mathcal{H}_i;
\varepsilon_i,
\mathcal{Z}_i
\right).
\label{eq:general_child}
\end{equation}
This editor may modify any harness component and is instructed to identify a broadly useful improvement from the task specification, current behavior, and evolutionary history. It therefore preserves the open-ended character of evolutionary search and can discover changes that are not associated with a single evaluator-defined failure.

For the specialized path, HGM-Dual selects two target errors from the observed error profile of $\mathcal{H}_i$. The two target error categories are selected as
\begin{equation}
c_{i,k}^{*}
=
\operatorname{TopK}_{c\in\mathcal{C}}
\left(
f_{i,c},\, k
\right),
\qquad
k\in\{1,2\},
\label{eq:target_error}
\end{equation}
where $f_{i,c}$ denotes the measured frequency of error category $c$ for $\mathcal{H}_i$. Here, $c_{i,1}^{*}$ and $c_{i,2}^{*}$ denote the first- and second-highest-ranked error categories, respectively.

For each selected target error, an error-conditioned child is generated as
\begin{equation}
\mathcal{H}^{\mathrm{err}}_{i,k}
=
\operatorname{Edit}_{\mathrm{err}}
\left(
\mathcal{H}_i;
c_{i,k}^{*},
\mathcal{S}_i^{c_{i,k}^{*}},
\mathcal{Z}_i
\right),
\qquad
k\in\{1,2\},
\label{eq:error_child}
\end{equation}
where $\mathcal{S}_i^{c_{i,k}^{*}}$ contains the traces and evaluator feedback associated with target error $c_{i,k}^{*}$. For each target, the specialized editor is asked to diagnose the corresponding failure, identify the responsible prompt, tool, or workflow behavior, and propose a focused correction while preserving capabilities that are already successful.

All three children are executed on the same optimization set and evaluated with the same reward function $\widehat{J}$. HGM-Dual retains
\begin{equation}
\mathcal{H}^{+}_i
=
\arg\max_{\mathcal{H}\in
\left\{
\mathcal{H}^{\mathrm{gen}}_i,
\mathcal{H}^{\mathrm{err}}_{i,1},
\mathcal{H}^{\mathrm{err}}_{i,2}
\right\}}
\widehat{J}(\mathcal{H}).
\label{eq:child_selection}
\end{equation}
The best child, its evaluation context, and edit attribution relative to $\mathcal{H}_i$ are added to the archive; the others are discarded. This dual-path procedure balances exploration and exploitation: the general child enables broad improvements, while specialized children target prominent failures. Retaining only the best child limits archive growth and focuses the search budget on promising harnesses.

\section{Experiments}

\begin{table}[t]
\centering
\caption{Performance comparison of task agent optimization for Travel and Shopping at an evaluation budget of 4000.   Best per column in \textbf{bold}, second best \underline{underlined}.}
\label{tab:results-4000}
\vspace{-2mm}
\setlength{\tabcolsep}{5pt}
\resizebox{.8\linewidth}{!}{%
\begin{tabular}{lccccc}
\toprule
& & \multicolumn{2}{c}{Travel} & \multicolumn{2}{c}{Shopping} \\
\cmidrule(lr){3-4} \cmidrule(lr){5-6}
Method & Evaluation Budget & Composite Score & Case Acc. (\%) & Composite Score & Case Acc. (\%)\\
\midrule
Seed Agent & 0 & 76.46 $\pm$ 0.22 & 25.00 & 81.91 $\pm$ 1.17 & 50.83 \\
\midrule
HGM      & 4000 & 77.06 $\pm$ 0.53 & 30.42 & 83.12 $\pm$ 0.27 & 50.83 \\
HGM-CA   & 4000 & 83.23 $\pm$ 2.65 & 54.17 & 85.32 $\pm$ 0.90 & 58.33 \\
HGM-CF   & 4000 & \underline{84.11} $\pm$ 0.07 & \underline{56.25} & \underline{86.67} $\pm$ 1.99 & \underline{60.83} \\
\textbf{HGM-Dual} & 4000 & \textbf{87.77} $\pm$ 1.31 & \textbf{64.38} & \textbf{89.23} $\pm$ 1.33 & \textbf{66.67} \\
\bottomrule
\end{tabular}
}
% \vspace{-5mm}
\end{table}

\textbf{Benchmark.}
We evaluate HGM-Dual on the \emph{DeepPlanning} benchmark~\cite{zhang2026deepplanning}, which consists of two constrained long-horizon planning tasks: travel planning and shopping. Both require maintaining global consistency across many interdependent decisions while satisfying numerous explicit and implicit constraints. In travel planning, the agent uses nine APIs to construct minute-level itineraries of up to seven days, evaluated by 21 feasibility checks across eight dimensions such as temporal, spatial, and budget consistency. In shopping, the agent searches an average of 171 records through 15 APIs over 45 product and user attributes, while jointly selecting products and coupons subject to product, eligibility, threshold, stacking, delivery, and budget constraints. These interacting constraints make locally valid decisions insufficient: the agent must ensure that the complete itinerary or cart remains globally feasible.

\textbf{Evaluation Metrics.} We assess each method along two dimensions. a) \emph{task performance}: For travel, we report a composite score that measures overall constraint satisfaction and case accuracy that measures the proportion of fully correct cases. For shopping, we report a match score that measures how many ground-truth products are correctly selected and case accuracy that measures exact cart-level matches. b) \emph{Compute efficiency}: Following HGM, we define the evaluation budget as the number of plans generated during optimization and apply this measure consistently to HGM, HGM variants, and HGM-Dual.

\textbf{Experimental Setup.} The optimization process starts from an initial harness of the task agent (seed agent) with a simple multi-turn loop, a minimal task prompt, and access to the task-specific tools of the benchmark. The agent editor is an LLM agent that updates the harness components (e.g., prompt, workflow, tool, verifier) of the task agent. The objective is to obtain the best-performing task agent by optimizing the harness alone; the underlying model weights are never updated. Information (e.g., code, logs, failing traces) supplied to the editor is gathered deterministically by a fixed procedure rather than through agentic exploration. For both travel planning and shopping tasks, we optimize on $60$ samples and evaluate on $120$ samples ($60$ unseen samples), so that generalization beyond the optimization set is measured explicitly. Unless otherwise specified, the agent editor uses GPT-5.4 (reasoning effort medium) and the task agent uses GPT-5.4-mini (reasoning effort high).

\paragraph{Error Categorization.} Error categories are produced by a deterministic, rule-based categorizer, one per domain. In travel, the categories follow the benchmark's evaluator: each of its eight commonsense dimensions (e.g., time feasibility, business hours) is a category, and violated hard constraints from the query are grouped by the entity they concern (flight, train, hotel, restaurant, attraction, or budget). Shopping considers four product level error categories (missing product, unmet feature, user-profile mismatch, not cheapest) and three cart level error categories (budget, cart not cheapest, coupon). 

\textbf{Baselines}. We compare HGM-Dual against ADAS \cite{hu2025automated}, GEPA \cite{agrawal2026gepa}, HGM \cite{wang2026huxley}, HGM-Customized (Agentic Editor) (HGM-CA)~\cite{zhang2026hyperagents}, and HGM-Customized (Fixed Context Editor) (HGM-CF). ADAS and GEPA are two LLM driven optimization process to improve agentic system performance, while HGM is a self-referential improvement algorithm in which a task agent edits its own implementation. Since our setting requires non-coding, long-horizon planning, we also consider HGM-CA (HGM-Customized (Agentic Editor)), a domain-customized variant of HGM. HGM-CA augments the original framework with domain-specific editing mechanisms designed to adapt the agent's prompts, tools, and workflow behavior to the planning setting. In this setting, the editor has agentic access with bash and file read/write tools to navigate task agents source code and logs. Similar to HGM-Dual's agent editor setting, we also consider HGM-CF (HGM-Customized (Fixed Context Editor)). In this setting, task agents code and logs are gathered deterministically and provided in the context of the agent editor to generate edits. HGM-CF gives HGM the same interface used by HGM-Dual, isolating the effect of agent evolution context and error-conditioned child generation. We do not compare with DGM \cite{zhang2026darwin} because HGM has previously been shown to outperform DGM in both task performance and improvement-budget efficiency.

\begin{figure}[t]
    \centering
    \includegraphics[width=\columnwidth]{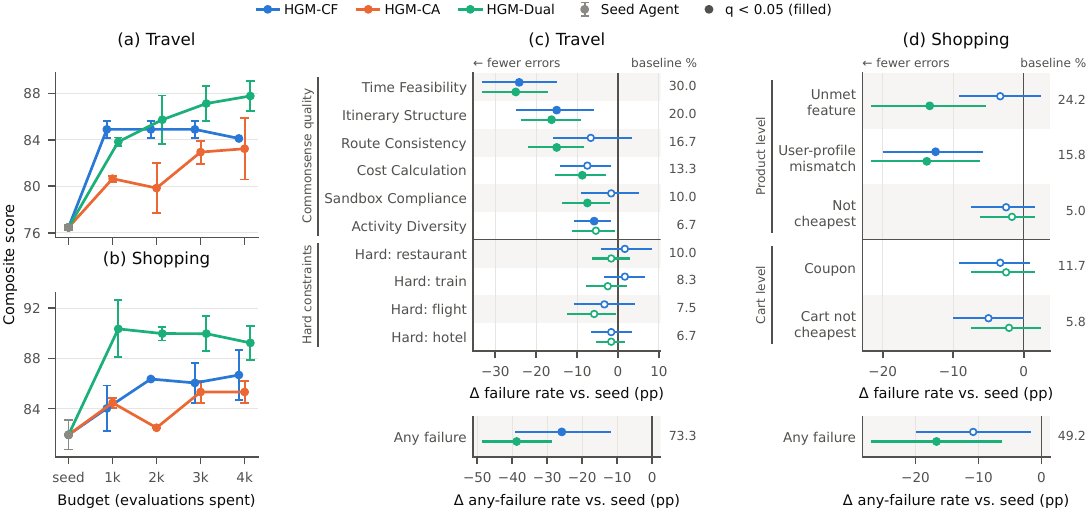}
    \caption{Budget efficiency and error breakdown. \textbf{(a,~b)} Best agent performance on \{1k, 2k, 3k, 4k\} evaluation budgets for travel and shopping task. \textbf{(c,~d)} Per-category change in failure rate of the best agent after optimization relative to the seed agent on the $60$ held-out task instances. Right-hand column gives the seed agent's failure rate, filled markers denote $q<0.05$.}
\label{fig:budget-errors}
\vspace{-3mm}
\end{figure}

\subsection{Evaluation}

\textbf{Performance Comparison to Baseline Approaches.} Table~\ref{tab:results-4000} reports the performance comparison to baselines (HGM, HGM-CA, HGM-CF) at $4000$ evaluation budget. Vanilla HGM barely moves beyond the seed harness on either domain, gaining only $0.60$ composite points on travel and $1.21$ on shopping, and leaving shopping case accuracy unchanged at $50.83$. This confirms that an editor designed for coding benchmarks transfers poorly to long-horizon planning, and motivates the customized variants. HGM-CF the strongest baseline on all four metrics ($84.11$ and $86.67$ composite, $56.25$ and $60.83$ case accuracy). HGM-Dual attains the best result on every metric in both domains, improving over HGM-CF by $3.66$ and $2.56$ composite points and by $8.13$ and $5.84$ case accuracy points on travel and shopping, respectively. Notably, the margin is consistently larger in case accuracy than in composite score, indicating that HGM-Dual does not merely accumulate partial credit across checkpoints but converts a larger fraction of plans into fully constraint-satisfying solutions

\textbf{Budget Efficiency.} We have analyzed the budget efficiency of HGM-Dual by evaluating the performance of selected best agents at different evaluation budget steps. We have considered selecting best agents at 1k, 2k, 3k and 4k evaluations and evaluate it on all $120$ instances for both travel and shopping task (Fig.~\ref{fig:budget-errors}(a,b)). On travel (Fig.~\ref{fig:budget-errors}(a)), HGM-CF is slightly ahead at 1k ($84.9$ vs.~$83.8$) but then stalls. HGM-CF returns the same agent at 1k, 2k and 3k and ends at $84.1$. HGM-Dual, in contrast, improves at every checkpoint, overtakes HGM-CF at 2k ($85.7$) and reaches $87.8$ at 4k. On shopping (Fig.~\ref{fig:budget-errors}(b)), HGM-Dual makes most of its gain within the first 1k evaluations, rising from $81.9$ to $90.4$, and stays near that level through 4k ($89.2$), whereas HGM-CF improves gradually and ends at $86.7$. The domain-customized variant HGM-CA remains below both methods across the budget range on travel and below HGM-Dual throughout on shopping. Overall, HGM-Dual converts the evaluation budget into performance far more efficiently than HGM-CF: it either keeps improving where HGM-CF plateaus (travel) or reaches HGM-CF's final score using roughly a quarter of the budget (shopping). Note that HGM-Dual's rejected child's evaluation is also counted towards the total evaluation budget used.

\textbf{Error Profile Analysis.} To identify the source of these gains, we decompose the failures of HGM-Dual and HGM-CF's final agent ($4$k budget) on the $60$ held-out task instances (Fig.~\ref{fig:budget-errors}(c,d)). Each row is an evaluator-defined error category, and each marker gives a method's change in the fraction of tasks failing that category relative to the seed agent; negative values indicate fewer errors. The right-hand column reports the seed agent's own failure rate. Rows are grouped into commonsense-quality and hard-constraint errors for travel and into product- and cart-level errors for shopping. Categories failed by the seed agent on fewer than $5\%$ of tasks are omitted, filled markers denote statistically significant changes, and the bottom panels report the share of tasks with any error. On travel (Fig.~\ref{fig:budget-errors}(c)), both methods eliminate most time feasibility and itinerary structure errors, the two most frequent categories. However, HGM-Dual is the only method that also removes route consistency errors (from $16.7\%$ of tasks to $1.7\%$, versus $10.0\%$ for HGM-CF) and sandbox compliance errors (from $10.0\%$ to $2.5\%$, versus $8.3\%$), and it reduces the share of tasks with any error from $73.3\%$ to $34.6\%$, compared with $47.5\%$ for HGM-CF. On shopping (Fig.~\ref{fig:budget-errors}(d)), the gap is concentrated in the most frequent category, unmet features: HGM-Dual more than halves it (from $24.2\%$ to $10.8\%$) whereas HGM-CF leaves it essentially unchanged ($20.8\%$). \emph{The advantage of HGM-Dual therefore stems more from resolving frequently appearing failure modes that HGM-CF leaves untouched.}

\begin{table}[t]
\centering
\caption{Performance comparison for Travel and Shopping at an evaluation budget of 1000.}
\vspace{-2mm}
\label{tab:results-1000}
\setlength{\tabcolsep}{5pt}
\resizebox{.8\linewidth}{!}{%
\begin{tabular}{lccccc}
\toprule
& & \multicolumn{2}{c}{Travel} & \multicolumn{2}{c}{Shopping} \\
\cmidrule(lr){3-4} \cmidrule(lr){5-6}
Method & Evaluation Budget & Composite Score & Case Acc. & Composite Score & Case Acc. \\
\midrule
Seed Agent & 0 & 76.46 $\pm$ 0.22 & 25.00 & 81.91 $\pm$ 1.17 & 50.83 \\
\midrule
ADAS     & 1000 & 80.83 $\pm$ 2.80 & 35.00 & 81.98 $\pm$ 2.15 & 51.25 \\
GEPA     & 1000 & 81.67 $\pm$ 0.66 & 39.58 & 81.18 $\pm$ 0.96 & 42.92 \\
HGM      & 1000 & 77.06 $\pm$ 0.53 & 30.42 & 83.12 $\pm$ 0.27 & 50.83 \\
HGM-CA   & 1000 & 80.65 $\pm$ 0.26 & 41.67 & \underline{84.46} $\pm$ 0.38 & 53.75 \\
HGM-CF   & 1000 & \textbf{84.90} $\pm$ 0.74 & \textbf{47.50} & 84.01 $\pm$ 1.83 & \underline{56.67} \\
\textbf{HGM-Dual} & 1000 & \underline{83.83} $\pm$ 0.37 & \underline{47.08} & \textbf{90.35} $\pm$ 2.27 & \textbf{70.83} \\
\bottomrule
\end{tabular}
}
% \vspace{-6mm}
\end{table}

\textbf{Performance Comparison at Low Budget Regime.} Due to the high cost of long-horizon agent evaluation and performance plateau, we evaluate ADAS and GEPA only at $1000$ evaluations. Table~\ref{tab:results-1000} reports the low-budget regime performance for different baselines. The two non-evolutionary optimizers improve over the Seed Agent on travel ($80.83$ and $81.67$ composite) but are essentially flat on shopping, where GEPA in fact falls below the seed harness in both metrics. Vanilla HGM is similarly ineffective at this budget. On Travel, the customized variant HGM-CF is the strongest method ($84.90$ composite, $47.50$ case accuracy), with HGM-Dual being second highest. On shopping, HGM-Dual dominates every baseline by a wide margin, reaching $90.35$ composite and $70.83$ case accuracy and exceeding the strongest competing method by $5.89$ and $14.16$ points respectively.

\paragraph{Impact of Agent Evolution Context.} We analyze and compare the edits made by two variants of HGM runs: HGM-CF and HGM-CF + Agent Evolution Context (no error-conditioned child generation). For each edit/expansion, we record its outcome against its parent on the cases both were evaluated on, its source diff, and the redundancy of its stated goal, measured as the maximum TF-IDF cosine against the goals of its ancestors. Table~\ref{tab:behavior-summary} shows that the Agent Evolution Context makes edits less redundant within a lineage ($0.22\rightarrow0.17$). Edits are less destructive: they worsen fewer instances ($37.6\%\rightarrow31.9\%$). Most importantly, adding Agent Evolution Context in the editor improves child agent's performance compared to their parent more often ($31.3\%\rightarrow41.0\%$). Therefore, the Agent Evolution Context alters the proposal distribution of the self-improvement steps positively rather than only improving the final score.

\paragraph{Impact of Error Conditioning.} We ablate error conditioning from HGM-Dual. In the ablated setting, the error-conditioned child agent generation prompt omits the error category but everything else remain the same as HGM-Dual (number of total children agents generated per expand, sequence of generation, and best of the child agent selection). At an evaluation budget of $1000$, the ablation scores 87.68 vs. 90.35 for HGM-Dual (Table~\ref{tab:uncond-benchmark}) attributing $~2\%$ improvement on the error conditioning. Our error analysis also show that error-conditioning results in resolving frequently appearing failure modes. Table~\ref{tab:uncond-benchmark} also shows adding evolution context on top HGM-CF boosts performance. Adding error conditional generation on top of HGM-CF + evolution context further improves performance.

% preamble: \usepackage{caption, booktabs}

\begin{table}[t]
\centering
\small

% ---------- Row 1 ----------
\begin{minipage}[t]{\linewidth}
\centering
\captionof{table}{Impact of the Agent Evolution Context on the optimization process (travel).}
\label{tab:behavior-summary}
\begin{tabular}{lcc}
\toprule
Edit characteristic & HGM-CF & \;+\,Agent Evolution Context \\
\midrule
Improves on its parent (\%) $\uparrow$        & 31.3 & \textbf{41.0} \\
Common instances made worse (\%) $\downarrow$ & 37.6 & \textbf{31.9} \\
Goal repeats from its lineage $\downarrow$    & 0.22 & \textbf{0.17} \\
\bottomrule
\end{tabular}
\end{minipage}
\vspace{.2em}

% ---------- Row 2 ----------
\begin{minipage}[t]{0.38\linewidth}
\centering
\captionof{table}{Impact of error conditioning.}
\label{tab:uncond-benchmark}
\begin{tabular}{lc}
\toprule
Method & Score \\
\midrule
HGM-CF                          & 84.01 $\pm$ 1.83 \\
\quad w/ Evolution context       & 88.21 $\pm$ 0.84 \\
HGM-Dual & 90.35 $\pm$ 2.27 \\
\quad w/o error cond. & 87.68 $\pm$ 0.29 \\
% \quad + Error cond.\ (HGM-Dual) & \textbf{90.35 $\pm$ 2.27} \\
\bottomrule
\end{tabular}
\end{minipage}
\hfill
\begin{minipage}[t]{0.58\linewidth}
\centering
\captionof{table}{Generalization to a different task-agent backbone (Mistral-Medium-3.5-128B).}
\label{tab:model_generalization}
\begin{tabular}{lcc}
\toprule
Method & Composite score & Case accuracy (\%) \\
\midrule
Seed Agent & 32 & \textbf{1.6} \\
HGM-CF        & 34 & 0.8 \\
HGM-Dual   & \textbf{42} & \textbf{1.6} \\
\bottomrule
\end{tabular}
\end{minipage}
\vspace{-10pt}
\end{table}

\textbf{Generalization Across Task Models.} To evaluate whether HGM-Dual generalizes across LLMs for task agent, we further conduct experiments with Mistral-Medium-3.5-128B~\cite{mistralmedium35}. In this setup, we still use GPT-5.4 (reasoning medium) for agent editor. As shown in Table~\ref{tab:model_generalization}, harness optimization improves the composite score for Mistral-Medium-3.5-128B, although its effect on case accuracy is limited. Standard HGM-CF provides only a modest increase in composite score, from \(32\) to \(34\), and reduces case accuracy from \(1.6\%\) to \(0.8\%\). In contrast, HGM-Dual raises the composite score to \(42\), outperforming the seed agent and HGM-CF by \(10\) and \(8\) points, respectively, while retaining the seed agent's case accuracy of \(1.6\%\).

\section{Conclusion}
We introduced HGM-Dual, a diagnostic-guided framework for efficient agent harness optimization. By incorporating structured agent-evolution context, HGM-Dual enables the editor to learn from prior edit outcomes and avoid redundant or ineffective modifications. Its error-conditioned dual-path generation further combines open-ended exploration with targeted refinement of error-relevant harness components. Experiments on travel and shopping planning tasks demonstrate that these complementary mechanisms outperform strong self-improving baselines while substantially reducing the required evaluation budget. Our ablations further show that evolution context provides consistent performance gains, whereas dual-path generation accelerates improvement under constrained budgets. Together, these results demonstrate that explicitly modeling evolutionary history and balancing general with error-targeted search can make agent harness evolution more effective and resource-efficient. Future work may extend the evolution context beyond individual lineages to share useful edit-outcome evidence across the entire search tree. Future work could jointly learn an evolving context and dual-path manager to adapt evidence selection, editing strategies, and resource allocation throughout the search. Finally, evaluating HGM-Dual across broader task domains and online environments would help characterize the transferability and long-term robustness of the evolved harnesses.

\section*{AI Use Statement}
In this work, all research ideas, methods, experiments, and findings are the work of the authors. We used generative AI tools for two tasks with recommended disclosure: (1) drafting initial versions of text throughout all sections of the paper, based on detailed outlines, notes, written descriptions, and experimental results provided by the authors; and (2) editing the paper to improve grammar, clarity, and readability. The methodology itself was designed and implemented entirely by the authors; AI tools were used only to help turn the authors' written descriptions of it into prose. We have reviewed all AI-assisted work: the authors revised all AI-drafted text, checked every technical statement, number, and citation against our own results and the original sources, and verified that the final text accurately reflects our work. We take responsibility for the final content of this work, including text, claims, or artifacts produced with the aid of generative AI.

\bibliographystyle{iclr2027_conference}
\bibliography{IEEEexample}

\appendix
\section{Appendix}
\subsection{Edit redundancy}\label{sec:edit_redundancy}
\paragraph{Edit-redundancy calculation for the shopping task.}
We analyze every non-root edit proposal in the search-tree archives of the
shopping task, yielding $N_{\mathrm{HGM}}=67$ proposals and
$N_{\mathrm{Dual}}=51$ proposals.

Each proposal is tagged using the following seven error categories:
\begin{enumerate}\label{eqn:error_cate}
    \item \textbf{Coupon correctness}: selecting, validating, or applying coupons correctly;
    \item \textbf{Profile compliance}: satisfying user-profile requirements such as gender or size;
    \item \textbf{Shipping constraints}: satisfying destination and delivery-time requirements;
    \item \textbf{Budget compliance}: respecting minimum or maximum spending constraints;
    \item \textbf{Cart completeness}: including all requested items and excluding unwanted items;
    \item \textbf{Exact feature matching}: satisfying explicit product attributes such as brand,
          color, season, rating, or review distribution; and
    \item \textbf{Price optimality}: selecting the cheapest valid products.
\end{enumerate}

The tags are determined from each proposal's optimization goal and proposed
changes using the same rules for both methods. Because an edit can address
multiple errors, the categories are not mutually exclusive. We represent
each proposal by the exact combination of its assigned categories, which we
refer to as its \emph{edit pattern}. For example,
\[
  (\text{profile compliance},\,
   \text{shipping constraints},\,
   \text{exact feature matching},\,
   \text{price optimality})
\]
is one edit pattern, whereas
\[
  (\text{shipping constraints},\,
   \text{exact feature matching})
\]
is another. Two proposals are considered to have the same edit pattern only
when their complete sets of error categories are identical.

For method $m$, let $n_m(s)$ be the number of proposals assigned to edit
pattern $s$. Its share of the tree archive is
\begin{equation}
    p_m(s) = 100 \times \frac{n_m(s)}{N_m}.
\end{equation}
The edit patterns are ranked independently for HGM and HGM-Dual in descending
order of $n_m(s)$. The figure reports the share of repeated edit pattern above. 

\subsection{Edit-targeting mismatch}

We treat each non-initial candidate as an edit and compare it with the candidate from which it was derived. Each edit's stated objective and proposed changes are mapped to functional error categories, using labels mentioned in Section \ref{eqn:error_cate}.

An edit is classified as having \emph{no corresponding parent error category} when it targets at least one functional component for which no error was reported in its parent. The metric is the fraction of edits satisfying this condition. The \emph{dominant error category} is the category associated with the largest number of failing evaluation cases. 

\subsection{Estimated Cost}
As shown in Table~\ref{tab:cost}, evaluating long-horizon planning agents
is computationally expensive, particularly after harness optimization, where
1k case evaluations can cost hundreds of dollars in task-agent API usage.
Given this cost, we prioritize the available evaluation budget toward
independent runs of the HGM variants, which are the primary focus of our
study. We therefore evaluate ADAS and GEPA at the 1k budget as reference
baselines, while reserving the more expensive 4k experiments for HGM,
HGM variants, and HGM-Dual.
\begin{table}[t]
\centering
\caption{Estimated task-agent API cost (USD) of 1000 samples evaluations on Travel and Shopping.
The task agent is GPT-5.4-mini (high reasoning effort).
Optimized averaged candidate agents evaluated by HGM and HGM-Dual.
Lower bounds assume ${\sim}80\%$ of input tokens are served from the prompt cache; upper bounds assume no caching.}
\label{tab:cost}
\setlength{\tabcolsep}{6pt}
\begin{tabular}{lcc}
\toprule
& Unoptimized & Optimized \\
\midrule
Travel   & \$400--800 & \$520--1{,}010 \\
Shopping & \$140--360 & \$300--910 \\
\bottomrule
\end{tabular}
\end{table}

\subsection{Full HGM-Dual}\label{sec:full-hgm-dual}
% ===================== Algorithm 1: full HGM-Dual =====================
Algorithm~\ref{alg:hgm-dual} summarizes the complete procedure.
HGM-Dual keeps the search loop of HGM~\citep{wang2026huxley}: at each
iteration, it either \emph{expands} the archive by modifying a selected
parent or \emph{evaluates} an existing harness on one additional task.
Both choices are made by Thompson sampling, and at the end of the search
the harness with the highest posterior success rate is returned. The only
change is in the expansion step. Instead of producing a single
open-ended modification, HGM-Dual calls \textsc{Expand}
(Algorithm~\ref{alg:hgm-dual-expansion}), which builds the evolution
context from the parent's lineage (Section~\ref{sec:agent_evolution_context}),
generates one general and two error-conditioned children
(Section~\ref{sec:dual_path_generation}), and adds only the strongest child to the
archive.

The symbols in Algorithm~\ref{alg:hgm-dual} are as follows. For each
harness $\mathcal{H}$, $n_{\mathrm{success}}(\mathcal{H})$ and
$n_{\mathrm{failure}}(\mathcal{H})$ count its passed and failed
evaluation tasks, while $n^{C}_{\mathrm{success}}(\mathcal{H})$ and
$n^{C}_{\mathrm{failure}}(\mathcal{H})$ aggregate these counts over its
clade, i.e., the subtree rooted at $\mathcal{H}$. The clade counts
measure how productive the descendants of $\mathcal{H}$ have been and
guide parent selection. The individual counts measure the harness's own
performance and guide evaluation and final selection. $N_t$ is the number
of evaluations spent so far. The archive is expanded when
$|\mathcal{A}| \le N_t^{\alpha}$, where $\alpha \in [0,1]$ controls how
quickly the tree widens relative to the evaluation budget. $\tau$ is a
scheduler that increases over the search. It sharpens the sampled Beta
distributions over time, shifting the search from exploration toward
exploitation. $S^{C}(\mathcal{H})$ and $S(\mathcal{H})$ are the
Thompson samples drawn from the clade-level and individual Beta
posteriors, used to select the parent to expand and the harness to
evaluate, respectively. Finally, $I_{\rho}$ is the $\rho$-percentile of
a harness's Beta posterior. Returning the harness that maximizes it
favors candidates whose high success rate is supported by sufficient
evaluations.
\begin{algorithm}[t]
\caption{Huxley--G\"odel Machine Dual (HGM-Dual)}
\label{alg:hgm-dual}
\begin{algorithmic}[1]

\REQUIRE Initial harness $\mathcal{H}_0$, widening parameter $\alpha$,
scheduler $\tau$, percentile $\rho$ for final selection,
error categories $\mathcal{C}$ with weights $w_c$.

\STATE Initialize archive $\mathcal{A}$ as a tree with root $\mathcal{H}_0$
\STATE Evaluate $\mathcal{H}_0$ and construct its evaluation context $\varepsilon_0$
\STATE Initialize counters $n^{C}_{\mathrm{success}}(\mathcal{H})$, $n^{C}_{\mathrm{failure}}(\mathcal{H})$,
$n_{\mathrm{success}}(\mathcal{H})$, $n_{\mathrm{failure}}(\mathcal{H})$ for all $\mathcal{H} \in \mathcal{A}$

\WHILE{evaluation budget is not exhausted}
  \IF{$|\mathcal{A}| \le N_t^{\alpha}$ \textbf{and} expandable parents exist}
    \STATE \textbf{Expand:}
    \FOR{each $\mathcal{H} \in \mathcal{A}$}
      \STATE Sample $S^{C}(\mathcal{H}) \sim \mathrm{Beta}\big(\tau(1+n^{C}_{\mathrm{success}}(\mathcal{H})),\, \tau(1+n^{C}_{\mathrm{failure}}(\mathcal{H}))\big)$
    \ENDFOR
    \STATE Select parent $\mathcal{H}_i = \arg\max_{\mathcal{H}} S^{C}(\mathcal{H})$
    \STATE $(\mathcal{A}, \mathcal{H}_i^{+}) \leftarrow \textsc{Expand}(\mathcal{H}_i, \varepsilon_i, \mathcal{A})$
    \hfill $\triangleright$ Algorithm~\ref{alg:hgm-dual-expansion}
    \STATE Update $n_{\mathrm{success}}, n_{\mathrm{failure}}$ for $\mathcal{H}_i^{+}$ from its selection evaluations
    \STATE Update $n^{C}_{\mathrm{success}}, n^{C}_{\mathrm{failure}}$ for $\mathcal{H}_i^{+}$ and its ancestors
  \ELSE
    \STATE \textbf{Evaluate:}
    \FOR{each $\mathcal{H} \in \mathcal{A}$ with remaining tasks}
      \STATE Sample $S(\mathcal{H}) \sim \mathrm{Beta}\big(\tau(1+n_{\mathrm{success}}(\mathcal{H})),\, \tau(1+n_{\mathrm{failure}}(\mathcal{H}))\big)$
    \ENDFOR
    \STATE Select harness $\mathcal{H}^{\star} = \arg\max_{\mathcal{H}} S(\mathcal{H})$
    \STATE Allocate a task to $\mathcal{H}^{\star}$ and update its evaluation context $\varepsilon^{\star}$
    \STATE Update $n_{\mathrm{success}}, n_{\mathrm{failure}}$ for $\mathcal{H}^{\star}$
    \STATE Update $n^{C}_{\mathrm{success}}, n^{C}_{\mathrm{failure}}$ for $\mathcal{H}^{\star}$ and its ancestors
  \ENDIF
\ENDWHILE

\RETURN $\arg\max_{\mathcal{H}\in\mathcal{A}}
I_{\rho}\big(1+n_{\mathrm{success}}(\mathcal{H}),\, 1+n_{\mathrm{failure}}(\mathcal{H})\big)$

\end{algorithmic}
\end{algorithm}

\subsection{Agent Evolution Context}\label{sec:appendix-evolution-context}

% Requires in the main preamble: \usepackage[most]{tcolorbox}
\definecolor{ctxframe}{HTML}{3C3C3C}
\definecolor{ctxback}{HTML}{F7F7F7}
\newtcolorbox{contextbox}[1]{%
  enhanced, breakable,
  colback=ctxback, colframe=ctxframe, coltitle=white,
  fonttitle=\bfseries, boxrule=0.7pt, arc=2pt,
  left=6pt, right=6pt, top=4pt, bottom=4pt,
  title={#1}%
}
% \cb inserts a zero-width breakpoint inside long typewriter identifiers,
% which TeX cannot otherwise hyphenate.
\newcommand{\cb}{\discretionary{}{}{}}
% Compact bullet list, defined without enumitem so no extra package is needed.
\newenvironment{ctxlist}{%
  \begin{list}{\textbullet}{%
    \setlength{\leftmargin}{1.2em}%
    \setlength{\labelwidth}{0.8em}%
    \setlength{\labelsep}{0.4em}%
    \setlength{\itemindent}{0pt}%
    \setlength{\listparindent}{0pt}%
    \setlength{\topsep}{2pt}%
    \setlength{\partopsep}{0pt}%
    \setlength{\itemsep}{1pt}%
    \setlength{\parsep}{0pt}%
  }%
}{\end{list}}
% Matching compact numbered list.
\newcounter{ctxenumi}
\newenvironment{ctxenum}{%
  \begin{list}{\arabic{ctxenumi}.}{%
    \usecounter{ctxenumi}%
    \setlength{\leftmargin}{1.6em}%
    \setlength{\labelwidth}{1.2em}%
    \setlength{\labelsep}{0.4em}%
    \setlength{\itemindent}{0pt}%
    \setlength{\listparindent}{0pt}%
    \setlength{\topsep}{2pt}%
    \setlength{\partopsep}{0pt}%
    \setlength{\itemsep}{1pt}%
    \setlength{\parsep}{0pt}%
  }%
}{\end{list}}

Each non-root node of the search tree carries an evolution record $\zeta_i$ that is summarized into the lineage context $Z_{p_r}$ supplied to the editor. We have provided one example from travel and shopping task each. 

\begin{contextbox}{Example Agent Evolution Context (Travel; one node of a lineage)}
\footnotesize
\setlist[itemize]{leftmargin=1.2em, topsep=2pt, itemsep=1pt, parsep=0pt}

\textbf{What was added}
\begin{itemize}
\item \texttt{select\_intercity\_transport} --- parses train/flight tool results, filters/sorts options, and returns one recommended intercity leg.
\item \texttt{\_trace} --- logs \texttt{platform\_core.trace} events.
\item \texttt{\_pick} --- returns the first non-empty value among alternate dict keys.
\item \texttt{\_parse\_json\_list} --- safely converts a JSON string into a list of dicts.
\item \texttt{\_route\_segments} --- extracts/orders \texttt{"Segment N"} route entries.
\item \texttt{workflow.py} transfer-line rule --- prompt now requires every transfer day to explicitly include the selected intercity transport line.
\item \texttt{tools\_schema.json} --- exposes \texttt{select\_intercity\_transport} as a callable tool.
\end{itemize}

\textbf{How it behaved (observed)}
\begin{itemize}
\item \texttt{mutable\_log} was present. Cumulatively: \texttt{decision\_branch/selector\_mode\_chosen} fired 122 times; \texttt{verifier\_fired} fired 216 times (\texttt{selector\_raw\_transport\_parsed} 122, \texttt{selector\_filters\_applied} 47, \texttt{selector\_recommended\_option} 47).
\item \texttt{select\_intercity\_transport} was used heavily: 122 calls, 0 errors observed. It co-occurred with all 26 passing and all 34 failing cases seen so far, so usage alone did not distinguish success from failure.
\item Instrumentation verdicts still did not track case outcome reliably: this batch had failing case \texttt{110} with \texttt{verifier\_fired/pass}, while passing cases \texttt{100}, \texttt{36}, \texttt{44}, \texttt{54}, and \texttt{57} had \texttt{verifier\_fired/fail}.
\item A concrete parse failure could coexist with success: case \texttt{54} had \texttt{selector\_raw\_transport\_parsed} fail (\texttt{mode=train}, \texttt{route\_count=0}), yet the case passed.
\end{itemize}

\textbf{What helped}
\begin{itemize}
\item The selector remained compatible with successful runs; this batch passed on \texttt{36}, \texttt{44}, \texttt{54}, \texttt{57}, \texttt{100}, and \texttt{107} with the tool active.
\item Some transport-focused cases did succeed, so the new selector and prompt rule did not broadly break itinerary generation.
\end{itemize}

\textbf{What didn't help (or hurt)}
\begin{itemize}
\item Transfer/time issues persisted: \texttt{10} and \texttt{110} failed \texttt{reasonable\_transfer\_time}; \texttt{33} failed \texttt{seamless\_intercity\_transfers}.
\item Transport-specific correctness still failed in some cases: \texttt{33} (\texttt{train\_seat\_status}) and \texttt{73} (\texttt{train\_cheapest\_train\_type}).
\item Non-transport failures also remained: accommodation traceability (\texttt{0}, \texttt{15}), restaurant/cuisine matching (\texttt{10}), and budget calculation (\texttt{33}).
\end{itemize}
\end{contextbox}

\begin{contextbox}{Example Agent Evolution Context (Shopping; one node of a lineage)}
\footnotesize
\emergencystretch=1.5em\relax

\textbf{What was added}
\begin{ctxlist}
\item \texttt{\_reset\_cart} --- clears the cart at \texttt{run\_task} start via public cart-mutation tools.
\item \texttt{\_build\_planning\_context} --- prefetches context and injects a level-specific planning checklist before phase 1.
\item \texttt{\_build\_verification\_message} --- builds a stronger phase-2 audit prompt from a cart snapshot plus a level-specific verification checklist.
\item \texttt{\_load\_cart\_json(wrapper)} --- returns the final cart state via \texttt{get\_cart\_info}.
\item \texttt{context\_prefetch/\cb user\_info\_prefetch} and \texttt{context\_prefetch/\cb cart\_snapshot\_\cb for\_verification} --- instrumentation for the two prefetch steps.
\item \texttt{decision\_branch/\cb level\_\{1,2,3\}\_\cb planning\_checklist} and \texttt{decision\_branch/\cb level\_\{1,2,3\}\_\cb verification\_checklist} --- instrumentation that the level-specific checklists were added.
\item \texttt{verifier\_fired/\cb tool\_reset\_cart\_\cb already\_empty} --- instrumentation that reset found nothing to delete.
\end{ctxlist}

\textbf{How it behaved (observed)}
\begin{ctxlist}
\item Cumulative observed performance is now \textbf{37/60} (prior record: 29/48; this batch: 8/12).
\item In this batch, both \texttt{context\_prefetch} events and both \texttt{decision\_branch} checkpoints fired in \textbf{all 12 cases} (8 pass, 4 fail). Combined with prior evidence, these checklist and prefetch paths have appeared in \textbf{all 60 observed cases}.
\item \texttt{verifier\_fired/\cb tool\_reset\_cart\_\cb already\_empty} logged \texttt{skip} in all 12 new cases; across observed batches, reset still has no logged deletion path.
\item The changed read/prefetch flow was definitely exercised: \texttt{get\_user\_info} was called in 12 cases and \texttt{get\_cart\_info} heavily (70 calls), each spanning both passing and failing cases.
\end{ctxlist}

\textbf{What helped}
\begin{ctxlist}
\item New passing cases: \texttt{L1-43}, \texttt{L1-45}, \texttt{L1-47}, \texttt{L2-29}, \texttt{L2-49}, \texttt{L2-6}, \texttt{L3-3}, \texttt{L3-6}.
\item Core search and read tools remained error-free in this batch: \texttt{search\_products}, \texttt{get\_product\_details}, \texttt{get\_cart\_info}, \texttt{get\_user\_info}, and \texttt{calculate\_transport\_time}.
\end{ctxlist}

\textbf{What didn't help (or hurt)}
\begin{ctxlist}
\item The added checklist and prefetch instrumentation remains non-discriminative: it also fired in every failing case (\texttt{L1-23}, \texttt{L1-7}, \texttt{L2-11}, \texttt{L3-10}).
\item This batch's failures were mostly cart mismatch errors: product swaps and missing extras in \texttt{L1-23}, \texttt{L1-7}, \texttt{L2-11}, and coupon mismatch in \texttt{L3-10}.
\item \texttt{add\_coupon\_to\_cart} remained unstable: 11 calls with 4 errors, and coupon handling was the failing area in \texttt{L3-10}.
\end{ctxlist}
\end{contextbox}

\subsection{Agent Editor Instruction}\label{sec:appendix-editor-prompt}

The box below gives the instruction issued to the agent editor at every expansion. 

\begin{contextbox}{Instruction Template for the Agent Editor}
\footnotesize
\emergencystretch=1.5em\relax

\textbf{Role and objective.}
You are the self-improvement module of a self-evolving agent. Diagnose what to
change from the feedback and the current code, then call
\texttt{submit\_self\_\cb improvement}. First understand the task: read the agent's
system prompt in \texttt{workflow.py}, and (when provided below) the tool
implementations, database schema, and evaluation scoring code---together they
show what each tool does, what the data looks like, and how output is graded.
Target edits at the failures that most affect the score.

\textbf{Editable files.} You may only modify these files in the
\texttt{task\_agent} workspace:
\begin{ctxlist}
\item \texttt{tool\_wrapper.py}, \texttt{tools\_schema.json}, \texttt{workflow.py}
\item any \texttt{*.py} file under \texttt{mutable\_tools/}
\end{ctxlist}

\textbf{Hard rules.}
\begin{ctxenum}
\item \texttt{workflow.py} MUST define \texttt{def run\_task(task: Task) -\textgreater{} AgentOutput}. The single argument must be named \texttt{task} (the validator enforces this).
\item \texttt{workflow.py} may import only: \texttt{platform\_core.\cb llm\_wrapper.call\_llm}, \texttt{platform\_core.runner} (\texttt{Task}, \texttt{AgentOutput}), \texttt{tool\_wrapper}, plus stdlib.
\item \texttt{tool\_wrapper.py} may import only: \texttt{platform\_core.tools}, \texttt{mutable\_tools.*}, plus stdlib.
\item Mutable tools (\texttt{mutable\_tools/*.py}) may import only: \texttt{platform\_core.tools}, sibling \texttt{mutable\_tools.*}, plus stdlib.
\item Reach capabilities only via \texttt{platform\_core.tools}: \texttt{call\_tool(name, **kwargs)} for immutable tools, and \texttt{call\_mutable\_tool(name, **kwargs)} for \texttt{mutable\_tools/*}. Never invoke a mutable tool's \texttt{run()} directly---routing through \texttt{call\_mutable\_tool} keeps its calls recorded in the trace.
\item \texttt{tools\_schema.json}: every entry's \texttt{name} must be backed by either an immutable tool OR a \texttt{mutable\_tools/\cb \textless{}name\textgreater{}.py} file. No collisions between immutable and mutable names.
\item Instrument your edits for the behavior summarizer. When you add a verifier, helper, or decision branch in \texttt{workflow.py} or a \texttt{mutable\_tools/*.py} file, call \texttt{platform\_core.trace.log(\cb label='your\_label', verdict='pass'|'fail'|'skip', name='specific\_check\_name', **context)} at the decision point. Conventions: pick a stable \texttt{label} (e.g.\ \texttt{'verifier\_fired'}, \texttt{'decision\_branch'}); use \texttt{name} to disambiguate within a label; set \texttt{verdict} to \texttt{pass}, \texttt{fail}, or \texttt{skip}. The summarizer cross-tabulates these logs against case outcomes so that the next editor knows which of your additions helped, which did not, and why. Skip instrumentation only for trivial edits (renames, docstrings).
\end{ctxenum}

\textbf{Required output.}
Call \texttt{submit\_self\_improvement} with a one-line \texttt{optimization\_goal},
a \texttt{proposed\_changes} summary, a \texttt{rationale}, and the \texttt{files}
payload. Each file is the FULL replacement content---do not produce diffs. Omit
files you do not change.
\end{contextbox}

% \bibliography{iclr2027_conference}
% \bibliographystyle{iclr2027_conference}

\end{document}